\documentclass[1p]{elsarticle}

\usepackage[hidelinks]{hyperref}
\journal{Expert Systems with Applications}

\biboptions{authoryear}

\newcommand{\citeonline}[1]{\citeauthor{#1} (\citeyear{#1})}
\usepackage{enumitem}
\usepackage{makecell}
\usepackage{threeparttable}
\usepackage{multirow}
\usepackage{booktabs}
\usepackage{color,soul}

\newcommand{\hlequation}[1]{#1}
\renewcommand{\hl}[1]{#1}

\begin{document}

\begin{frontmatter}

\title{NASirt: AutoML based learning with instance-level\\complexity information}

%% Group authors per affiliation:
%\author{Habib Asseiss Neto\fnref{myfootnote}}
%\address{Instituto Federal de Mato Grosso do Sul, Três Lagoas, Brazil \texttt{habib.asseiss@ifms.edu.br}}
%\fntext[myfootnote]{Since 1880.}

%% or include affiliations in footnotes:
%\author[address1,address2]{Habib Asseiss Neto\corref{mycorrespondingauthor}}
%\cortext[mycorrespondingauthor]{Corresponding author}
%\ead{habib.asseiss@ifms.edu.br}
%
%\author[address3]{Ronnie Cley de Oliveira Alves}
%\author[address2]{Sérgio Vale Aguiar Campos}
%
%
%\address[address1]{Instituto Federal de Mato Grosso do Sul, Três Lagoas, Brazil}
%\address[address2]{Departamento de Ciência da Computação, Universidade Federal de Minas Gerais, Belo Horizonte, Brazil}
%\address[address3]{Instituto Tecnológico Vale, Belém, Brazil}

\author{Habib Asseiss Neto\corref{cor1}}
\ead{habib.asseiss@ifms.edu.br}
\address{
Federal Institute of Mato Grosso do Sul, Três Lagoas, MS, Brazil. \\
Department of Computer Science, Federal University of Minas Gerais, Belo Horizonte, MG, Brazil.
}

\author{Ronnie Cley de Oliveira Alves}
\ead{alvesrco@gmail.com}
\address{
Instituto Tecnológico Vale, Belém, PA, Brazil. \\
Federal University of Pará, Belém, PA, Brazil.
}

\author{Sérgio Vale Aguiar Campos}
\ead{scampos@dcc.ufmg.br}
\address{
Department of Computer Science, Federal University of Minas Gerais, Belo Horizonte, MG, Brazil.
}

\cortext[cor1]{Corresponding author}

\newpageafter{author}

\begin{abstract}
Designing adequate and precise neural architectures is a challenging task, often done by highly specialized personnel. AutoML is a machine learning field that aims to generate good performing models in an automated way. Spectral data such as those obtained from biological analysis have generally a lot of important information, and these data are specifically well suited to Convolutional Neural Networks (CNN) due to their image-like shape. In this work we present NASirt, an AutoML methodology based on Neural Architecture Search (NAS) that finds high accuracy CNN architectures for spectral datasets. The proposed methodology relies on the Item Response Theory (IRT) for obtaining characteristics from an instance level, such as discrimination and difficulty, and it is able to define a rank of top performing submodels. Several experiments are performed in order to demonstrate the methodology's performance with different spectral datasets. Accuracy results are compared to other benchmarks methods, such as a high performing, manually crafted CNN and the Auto-Keras AutoML tool. The results show that our method performs, in most cases, better than the benchmarks, achieving average accuracy as high as \hl{97.40\%}.
\end{abstract}


\end{frontmatter}

%\linenumbers

\section{Introduction}

Analysis of biological data is a very important task performed in order to obtain useful knowledge from a specific subject. A widely used method that can be applied for analyzing the structural composition of materials is called spectroscopy analysis. It allows extracting the material fingerprint through the emission of light in the sample \citep{Baker2016}. Spectral data obtained from spectroscopy analyses consist of data coordinates in the shape of a curve, called spectrum, that represents the analyzed material. From these data it is possible to perform different analyses that are common in a wide range of application domains such as food sciences, pharmaceutical, medicine, geology, and others \citep{Mukrimin2019, Gondim2017, Stables2017, Acquarelli2017, Liu2017}.

Machine learning techniques can assist in this task, and they provide ways of understanding and producing useful knowledge from spectral data. Specifically, Convolutional Neural Networks (CNNs) are well suited to this problem due to their high accuracy in pattern recognition. In fact, several studies have shown that CNN architectures achieve high accuracy on spectrum-like data structures \citep{Liu2017, Makantasis2015, Berisha2019}. However, there are a multitude of forms that CNNs can be constructed and their architecture can become complex \citep{Elsken2019}. Typically, the process of finding a good neural architecture depends on manual analyses \citep{Bergstra2012}. Recently, Automated Machine Learning (AutoML) has become a popular subfield of machine learning that focuses on automating the process of selecting the right neural architectures, training procedures, and hyperparameters optimization to a specific problem \citep{Hutter2019}. One AutoML approach is Neural Architecture Search (NAS), which aims at obtaining automated architectures of neural networks that achieve high predictive performance on unseen data \citep{Elsken2019}.

It has been recently demonstrated that analyzing dataset instances complexities can improve the learning performance of machine learning methods \citep{Smith2014}. The work from \citeonline{MartinezPlumed2019}, uses concepts from Item Response Theory (IRT), a classic psychometrics approach for evaluating tests and subjects, in order to determine instance difficulty, discrimination, and other parameters. To relate IRT and machine learning, the test subjects are considered machine learning models, which respond to test items. These items are the instances of a dataset, and the given responses are the models' predictions. The use of IRT allows the estimation of a machine learning model's ability regarding the instances that it is submitted to \citep{MartinezPlumed2019}.

In this work, we propose NASirt: a methodology based on NAS and IRT that can determine complexities from instances belonging to spectral datasets and it is able to generate adequate CNN architectures to perform high accuracy predictions. The methodology can also give an overview of the hyperparameters that generate models with higher abilities. The proposed method is considered a machine learning meta-model, or a meta-learning technique, since it is dependent on different CNN models. We have used the meta-model to perform several experiments with different spectral datasets, analyzing interesting outputs such as instance difficulty and discrimination distributions, models' abilities and classification accuracies. Also, we have analyzed the complexity of the generated models by counting the total number of network parameters, and when compared to other methods, it has shown to generate low complexity models. Finally, we have compared the results with a benchmark model and other techniques, and the results show that the proposed methodology is able to generate, on average, higher accuracy on all experiments performed.

Recently, AutoML and NAS studies have gained attention, and some approaches have outperformed manually designed network architectures on different tasks like image classification, object detection and semantic segmentation \citep{Elsken2019}. IRT, on the other hand, is a well-established theory, used for decades in psychometrics. However, only in the last few years its application on machine learning was proposed. Some studies have focused on the analysis of general machine learning classifications \citep{MartinezPlumed2016, MartinezPlumed2019}, while some works have focused on specific Natural Language Processing methods \citep{Lalor2016, Lalor2018}. This study explores the utilization of IRT estimators to provide the building blocks to automate the design of neural architecture machines, specifically for classification problems and datasets formed by spectral analyses. The rest of the paper is organized as follows. Section~\ref{sec:materials_and_methods} presents a brief overview of CNNs and IRT and introduces the proposed  AutoML methodology. Section~\ref{sec:experiments} details the datasets and some experiments comparing our method to others. Finally, Section~\ref{sec:conclusion} presents the conclusion of the paper and discusses possibilities for future works.

\section{Materials and methods}
\label{sec:materials_and_methods}

In this work we propose NASirt: a meta-learning methodology with a NAS based approach, that uses Convolutional Neural Networks and Item Response Theory, in order to have a better understanding of spectroscopy samples and the learning abilities of machine learning models. We propose a classification method that tries to find an architecture that maximizes performance on unseen data. To do that, we train several models with hyperparameter variations and submit their responses to an IRT model. Then, we have difficulty and discrimination parameters to test instances using the most adequate model. The procedure generates several interesting information that we analyze in this section.

\subsection{Background}

\subsubsection*{Convolutional Neural Networks}

Data obtained from spectroscopy processes are generally analyzed by statistical and computational models in order to reveal important information of the samples \citep{Acquarelli2017}. Convolutional Neural Networks (CNNs) have been successfully applied to different types of spectral data \citep{Baker2016, Min2017, AsseissNeto2019}. CNNs are widely and successfully applied in image recognition and computer vision problems \citep{Zeiler2014, Min2017}, but they also can recognize features from the spectrum data, without any additional data preprocessing \citep{Acquarelli2017, Liu2017}.

A general CNN architecture contains one or more convolutional layers together with non-linear activation layers and pooling layers. Convolutional layers consist of a set of filters that are computed during training and generate feature maps \citep{Liu2017}. Filters are applied repeatedly across the entire dataset, so patterns can be detected when identical regions in the data are found. This also improves training efficiency since the number of features to learn is reduced.  Then non-linear layers increase the non-linear properties of feature maps. Finally, in pooling layers, maximum or average subsampling of non-overlapping regions in feature maps is performed \citep{Min2017}.

Although CNNs are used with great predictive performance for several different applications \citep{Acquarelli2017}, there is no clear relationship of their architecture and hyperparameter configurations and specific abilities in correctly learning and predicting information \citep{Yosinski2015, Min2017}. Therefore, these machine learning models are considered ``black boxes'' \citep{Yosinski2015, Min2017}. Recent works have addressed the understanding of the internals of deep learning models \citep{Zeiler2014, Li2017}. Other works focus on models' prediction interpretations, detecting feature importance and output explanations \citep{Lundberg2017, Ribeiro2016}.

\subsubsection*{Item Response Theory}

Item Response Theory (IRT) is a widely used methodology in psychometrics that addresses the concepts behind tests, questionnaires, and similar instruments. It typically measures abilities and attitudes of a human individual considering the analysis of their responses about items in a questionnaire \citep{Baker2017}. The theory considers that every individual being examined that responds to a test item has an unobserved latent trait dimension, called ``ability''. IRT adopts a model for the probability of each possible response to an item and this probability is a function of the latent trait and some item characteristics. One of the most flexible IRT models is called \hl{3PL (the Three-Parameter Logistic) model}, which considers the following item parameters: discrimination ($a$), which describes how well an item can differentiate between individuals having high or low abilities; difficulty ($b$), which is the level of ability that produces a chance of correct response; and guessing ($c$), which is the probability that an individual \hl{will guess} an item correctly \citep{Baker2017, Lalor2016}. Lastly, the individual's general score for the questionnaire is known as the true score \citep{Baker2017}. 

IRT can be applied to the machine learning context by creating a relationship between individuals and models, and between test items and instances. The individual that responds the questionnaire are considered as the model inferred by training, while the items of a questionnaire are the instances that belong to a dataset. According to \citeonline{MartinezPlumed2019}, the proficiency (or ability) of a machine learning method can be understood as the difficulty level whose problems the method is able to solve.

% Using IRT and instance level information (difficulty, discrimination)

\subsection{Proposed methodology}
\label{subsec:proposed_methodology}

Based on NAS and IRT concepts, NASirt is a meta-learning methodology focused on supervised classification in which datasets consist of spectral data points, obtained from techniques based on spectroscopy. Since CNNs have shown to be particularly suited to these kinds of data structure, they are used as sub-models in our method. The methodology relies on independent training of CNN models based on the hold-out validation method. \hl{Initially, it is necessary} to define the desired dataset folds for training and test sets. In order to demonstrate the consistency of results, different folds are suggested, for instance, 90\% for training and 10\% for testing, 75\% for training and 25\% for testing, and 50\% for training and 50\% for testing. Instances in each fold are randomly selected from the original dataset. The following steps are repeated independently for each training/test fold.

\hl{The Step 1 of the methodology is to train a collection of CNN models with hyperparameter variations. We consider a set of hyperparameters to be explored in a grid-search manner, combining hyperparameters in order to generate $m$ different models. Such hyperparameters are the ones more commonly found in literature and practice for high performing CNN architectures \mbox{\citep{Liu2017, Ramirez2015, AsseissNeto2019}} and they include the number of convolutional filters, kernel size, dropout rate, dense layer size, etc. The specific hyperparameters and values used are presented below in Section~{\ref{sec:experiments}}.} Then, the test set instances are submitted to all $m$ trained models \hl{(Step 2)}, collecting each model's response to each instance. The responses are used to generate the IRT model, which provides important parameters about the dataset instances, such as difficulty and discrimination \hl{(Step 3)}. The true score for each model is also calculated. IRT coefficients are normalized between 0 and 1 \hl{(Step 4)}.

According to \citeonline{MartinezPlumed2019}, in order to avoid singularity in IRT model, i.e., in a possibility that all classifiers were right for very easy instances, it is recommended to introduce specific artificial classifiers. So, apart from the trained models, we introduce artificial classifiers to ensure the IRT model doesn't suffer from singularity in responses \citep{MartinezPlumed2019}. \hl{The method relies on three artificial classifiers, called ``random classifiers'', that output class prediction for instances randomly with equally distributed probabilities. Two special artificial classifiers are also used: the optimistic, which is a classifier that predicts all instances correctly, and the pessimistic, that always predict them incorrectly.}

The true score can be used to rank all trained models according to their abilities considering the test set. Models with higher true score generally leads to higher overall test accuracy \citep{MartinezPlumed2019}. Therefore, the next step in the methodology is to create a model rank with the highest $n$ true scores, where $n$ is a relatively low number of models, i.e., 5 or 10 models \hl{(Step 5)}. The selected $n$ models consist of models with higher abilities specific to the test instances, so it is safe to discard all the remaining models from the model collection.

Now instead of testing all models with all test instances, we can use the difficulty and discrimination information in order to separate instances for testing with the most adequate model. The model with the highest ability is more likely to perform better with more difficult instances, while less difficult instances can be submitted to models with lower abilities. The same idea can be applied to discrimination. 

So we define one of two IRT parameters: $b$ (difficulty) or $a$ (discrimination) and sort all instances from the test set according to the defined IRT parameter value. We separate instances into $n$ bins so that every bin has roughly the same number of instances \hl{(Step 6)}. Since instances are in ascending order, the first bin has instances with lower values of the selected IRT parameter (difficulty or discrimination), and every subsequent bin has more difficult (or discriminative) instances. Then, we perform the test with each of the $n$ models with instances from a specific bin, with the lowest score model being tested with instances from the first bin, the second lowest score model being tested with the second bin, and so on \hl{(Step 7)}.

\begin{figure}[ht!]
\centerline{\includegraphics[width=1\textwidth]{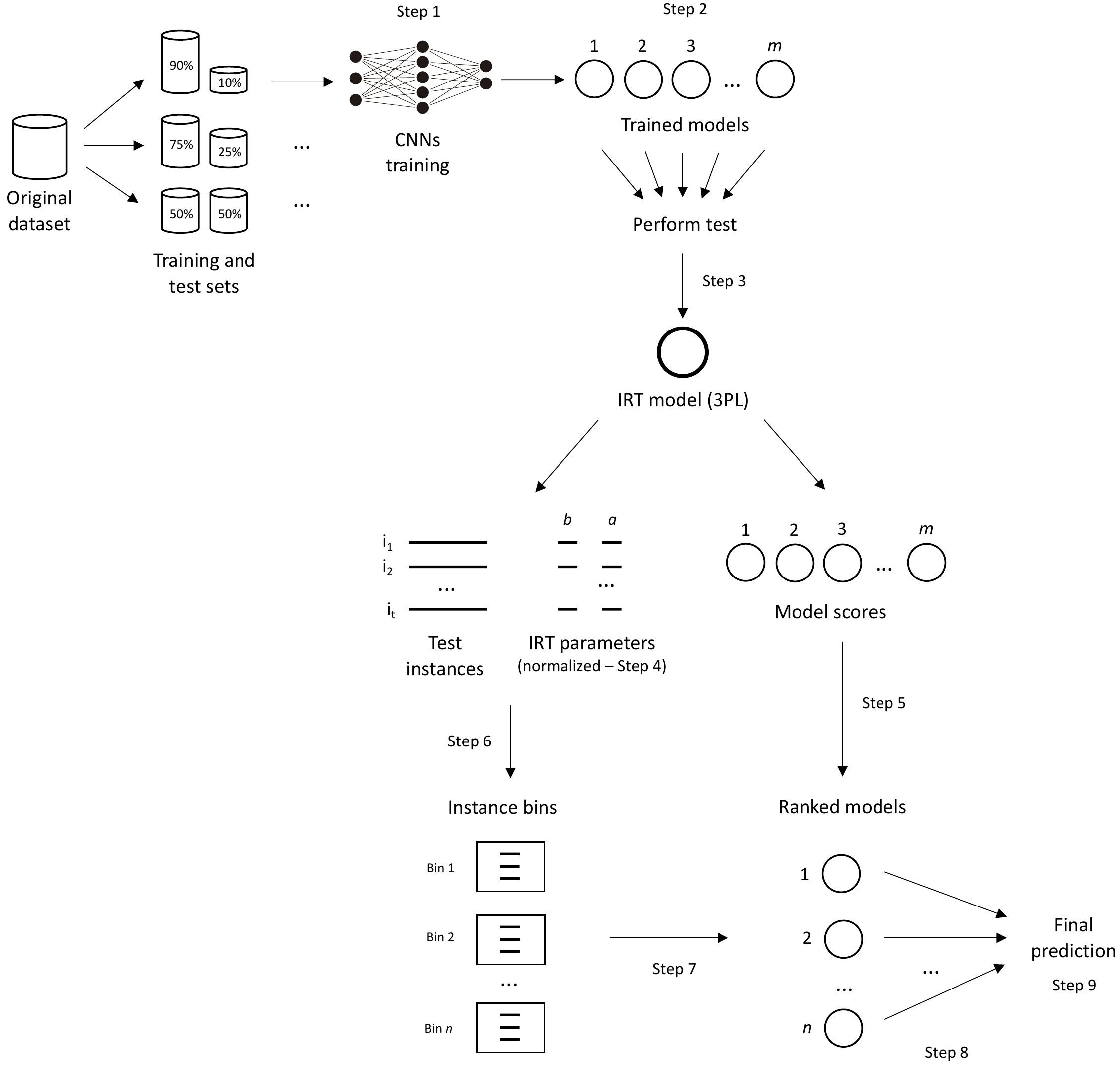}}
\caption{Overview of the proposed AutoML classification methodology.}
\label{fig:methodology}
\end{figure}

Finally, predictions performed in each bin are considered in order to calculate accuracy for each model. Then, we consolidate every classification to get general results considering all instances and all models utilized \hl{(Step 9)}. In this consolidation step, the final accuracy for the method is calculated simply by counting correct predictions in all bins, and dividing it by the total number of instances in the test dataset. That consolidation is the final prediction for the meta-learning classifier. The following steps briefly summarize the proposed method, for each training/test fold. Figure \ref{fig:methodology} shows a diagram of the main steps that are part of our method.

\setlist{nolistsep}
\begin{enumerate}[leftmargin=5em, label=Step \arabic*.]
	% \item Generate CNNs models with a predefined hyperparameter variation, obtaining $m$ models
	\item \hl{Train $m$ models using grid-search for all combination of hyperparameters defined }
	\item Submit all instances in the test set to all $m$ models
	\item \hl{Generate an IRT model using each model's instance predictions (considering them as item responses)}
	\item \hl{Normalize IRT parameters difficulty and discrimination}
	\item Select $n$ best models based on IRT true scores
	\item \hl{For each IRT parameter $b$ (difficulty) or $a$ (discrimination), separate test instances into $n$ difficulty or discrimination bins}
	\item Perform each $n$ model classification considering a specific bin, where higher ability models classify more difficult (or discriminative) instances
	\item Calculate accuracy for each model
	\item Consolidate accuracies
\end{enumerate}

\vspace{1em}

\hl{Formulas {\ref{eq:f1}}, {\ref{eq:f2}}, {\ref{eq:f3}}, and {\ref{eq:f4}} below describe higher level operations and give an overview of how the method works.}

\begin{equation}
\label{eq:f1}
\hlequation{
    \left \langle A, B, S \right \rangle = \textsc{IRT}(\mathbf{y}_1, \mathbf{y}_2, ..., \mathbf{y}_m)
}
\end{equation}
\begin{description}
\item \hl{where}
\item[] \hspace{0.8em} \hl{$\mathbf{y}_1, \mathbf{y}_2, ..., \mathbf{y}_m$ are the prediction vectors for the $m$ trained CNNs;}
\item[] \hspace{0.8em} \hl{\textsc{IRT} is the 3PL model, which takes instance predictions for each CNN;}
\item[] \hspace{0.8em} \hl{$A$ is the obtained instance discrimination vector;}
\item[] \hspace{0.8em} \hl{$B$ is the obtained instance difficulty vector;}
\item[] \hspace{0.8em} \hl{$S$ is the calculated score for each model.}
\end{description}

\begin{equation}
\label{eq:f2}
\hlequation{
    \left \langle M'_1, M'_2, ..., M'_n \right \rangle = \textsc{Rank}(\ (M_1, S_1), (M_2, S_2), ..., (M_m, S_m)\ )
}
\end{equation}
\begin{description}
\item \hl{where}
\item[] \hspace{0.8em} \hl{\textsc{Rank} is a function that returns the models with higher abilities;}
\item[] \hspace{0.8em} \hl{$M_1, M_2, ..., M_m$ are CNN models;}
\item[] \hspace{0.8em} \hl{$S_1, S_2, ..., S_m$ are IRT scores of the related model;}
\item[] \hspace{0.8em} \hl{$M'_1, M'_2, ..., M'_n$ are ranked CNN models with higher abilities.}
\end{description}

\begin{equation}
\label{eq:f3}
\hlequation{
    \left \langle bin_1, bin_2,...,bin_n \right \rangle = \textsc{SeparateBins}(\ (I_1, I_2,..., I_t), A, B\ )
}
\end{equation}
\begin{description}
\item \hl{where}
\item[] \hspace{0.8em} \hl{\textsc{SeparateBins} is a function that separates test set into instance bins;}
\item[] \hspace{0.8em} \hl{$I_1, I_2,..., I_t$ are instances from the test set;}
\item[] \hspace{0.8em} \hl{$A$ is the instance discrimination vector;}
\item[] \hspace{0.8em} \hl{$B$ is the instance difficulty vector.}
\end{description}

\begin{equation}
\label{eq:f4}
\hlequation{
    \textsc{NASirt}(M'_1, M'_2, ..., M'_n) = \textsc{Classify}(\ (bin_1, bin_2,...,bin_n), (M'_1, M'_2, ..., M'_n)\ )
}
\end{equation}
\begin{description}
\item \hl{where}
\item[] \hspace{0.8em} \hl{$bin_1, bin_2,...,bin_n$ are instance bins separated by discrimination or difficulty;}
\item[] \hspace{0.8em} \hl{$M'_1, M'_2, ..., M'_n$ are ranked CNN models;}
\item[] \hspace{0.8em} \hl{\textsc{Classify} is a function that submits instances in each bin to the appropriate model.}
\end{description}

\section{Experiments}
\label{sec:experiments}

\hl{We performed a series of experiments in order to demonstrate the characteristics of the proposed methodology. Each experiment is related to a dataset, and for each dataset, we used bootstrap sampling to repeatedly select three random runs. The bootstrap method can estimate performance in a more realistic way and also keep control of training and test set proportions. In each experiment}, we compare our technique to an available high performance single CNN model, to a majority-voting approach that combines classifications from all trained models, and also to the execution of an AutoML tool named Auto-Keras. The objective is to show that our method is capable of achieving at least similar results to the benchmark methods. Unlike the benchmark model's architecture, that is specifically crafted for a problem, and voting, that accounts predictions from all models, our method selects an adequate combination of hyperparameters in an automated way. Besides that, the proposed methodology brings additional information such as instance quality and model abilities, provided by the IRT approach. Hence even a similarity of results indicates that our method is preferred.

In addition to high performance classification capabilities, the proposed AutoML method produces some interesting support data related to explainability of the models and instance complexities. This information can be used to obtain a better understanding of the analyzed dataset and its instance characteristics. So, for each dataset we provide some additional information, besides the final classification accuracy, that are described in each following experiment.

The training of the initial CNN model collection considered the same hyperparameter variations in each experiment. The hyperparameters considered for training are the number of convolution layers and filters, the kernel size (length of the convolution window), the number of dense layers, the number of neurons on each dense layer, the dropout rate \citep{Srivastava2014}, the max-pooling window size, and the type of activation layers \citep{Maas2013}. CNN models were trained with every possible combination of values from these hyperparameters, presented in Table~\ref{tab:hyperparameters}. Some characteristics were kept fixed in order to generate the network architectures, like the presence of layer normalizations \citep{Ioffe2015}, the negative slope coefficient was set to 0.3 when the activation layer is LeakyReLU, and the Adam optimizer \citep{Kingma2014} was used in every combination. In each experiment related to one of the datasets, and for each training fold (90\%, 75\%, and 50\% of the original dataset), we generated 384 different models through the combination of all hyperparameters defined.

\begin{table}[ht!]
\centering
\caption{Hyperparameter values used to generate the CNN collection as the first step of the methodology.}
\label{tab:hyperparameters}
\resizebox{\textwidth}{!}{
\begin{tabular}{lp{8cm}l}
\toprule
Hyperparameter & Description & Possible values \\ 
\midrule

Convolution layers &
Number of convolution layers &
1, 2  \vspace{5pt}\\

Convolution filters &
Number of convolutional filters that learn input features &
8, 32, 128  \vspace{5pt}\\

Kernel size &
Convolution kernel size, i.e., the length of the convolution window &
8, 16  \vspace{5pt}\\

Dense layers &
Number of fully connected, regular neural network layers that follow the convolutional layers &
1, 2  \vspace{5pt}\\

Dense size &
Number of neurons that compose each fully connected layer &
128, 1024  \vspace{5pt}\\

Dropout rate &
Rate of randomly ignored neurons in order to prevent the network from overfitting &
0, 0.4  \vspace{5pt}\\

Max Pooling size &
The size of Max Pooling window, which downsamples the input representation by taking the maximum value over the window &
0, 4  \vspace{5pt}\\

Activation &
The activation function performed, i.e, the function that defines the output of a node given an input in a layer &
\texttt{leakyrelu}, \texttt{tanh} \\

\bottomrule
\end{tabular}
}
\end{table}

CNN architectures were implemented in Keras \citep{Chollet2015} and TensorFlow \citep{Abadi2015} in Python. The IRT analyses used the MIRT package in R \citep{Chalmers2012} and was integrated into the Python workflow using the `rpy2' library.

\subsection{Dataset selection}

In order to experiment with the proposed methodology, we have collected 3 different datasets containing spectral data from different sources. Dataset 1 is called Milk Adulterants and contains infrared spectra from thousands of pure and adulterated milk samples. Dataset 2 is called Milk Whey and also contains milk spectra, but they are added with different proportions of milk whey. Dataset 3 is called Trees, and contains spectral readings from different tissues from different tree species. Table \ref{tab:datasets} shows the number of instances, features and classes for each dataset, as well as class distributions. Since datasets refer to spectral information, the number of features refer to the number of data points, or coordinates, that form each spectrum, and it is related to the precision or resolution that the spectroscopy was performed on the samples.

\begin{table}[ht!]
\centering
\caption{Datasets used for experimenting our meta-learning method.}
\label{tab:datasets}
\resizebox{\textwidth}{!}{
\begin{tabular}{llllll}
\toprule
ID & Dataset & \# Instances & \# Features & \# Classes & Class Distribution \\ 
\midrule
1 & Milk Adulterants & 4846 & 518 & 6 & (50.1\%, 9.5\%, 10.0\%, 9.6\%, 9.9\%, 10\%) \\ 

2 & Milk Whey & 1040 & 518 & 2 & (50\%, 50\%) \\ 

3 & Trees & 1270 & 1154 & 4 & (20\%, 24.4\%, 28.1\%, 27.5\%) \\ 

\bottomrule
\end{tabular}
}
\end{table}

\subsubsection*{Dataset details}

\begin{figure}[ht!]
\centerline{\includegraphics[width=0.95\textwidth]{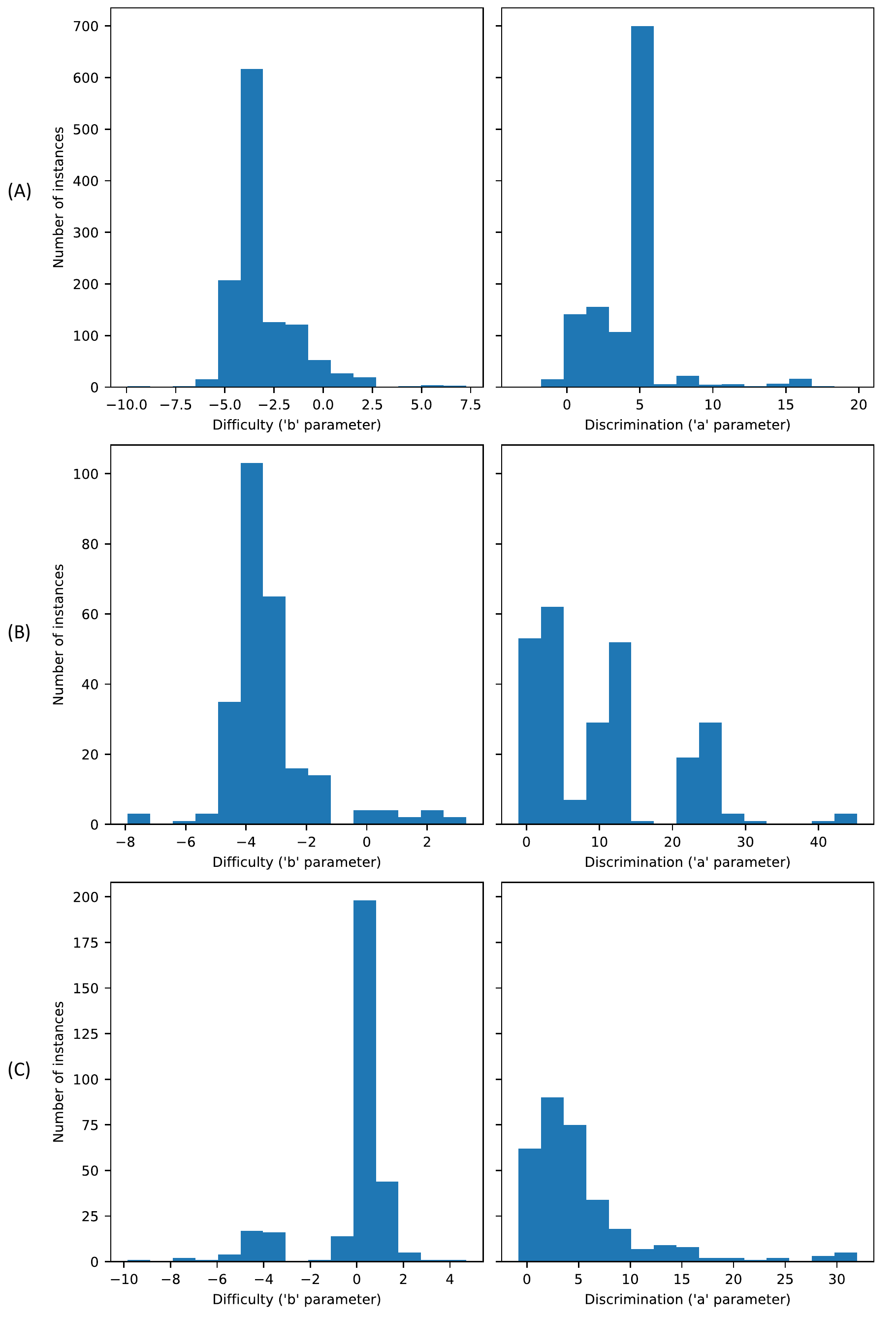}}
\caption{Difficulty and discrimination visualization of instances from the 75/25\% fold of (A) Milk Adulterants Dataset, (B) Whey Dataset, and (C) Trees Dataset.}
\label{fig:instances_b_a}
\end{figure}

Datasets 1 and 2 are composed of milk samples provided by the Laboratory for Milk Quality Analysis (Accredited ISO/IEC 17025) from the Federal University of Minas Gerais, Brazil \citep{AsseissNeto2019}. In Dataset 1, nearly half of the samples were adulterated with one of five known substances (bicarbonate, formaldehyde, peroxide, starch and sucrose), so it consists of either pure or adulterated milk. The addition of substances replicated the characteristics of fraud commonly found in the dairy industry. Adulterants were diluted in milk and submitted to a Fourier-Transformed Infrared (FTIR) spectroscopy equipment, that outputted sample spectra. In the first experiment, using Dataset 1, we have performed the classification of instances within 6 classes: ``raw'' or one of the 5 substances. IRT modeling allows the determination of difficulty and discrimination parameters of test instances from the dataset, which are shown in Figure~\ref{fig:instances_b_a}-(A).

Dataset 2 was developed in a similar manner as Dataset 1, but it contains samples from different sources in the same laboratory. Nearly half the samples were added extra amounts of milk whey, while the other half was kept as pure milk. Adding milk whey is considered a different kind of adulteration, since it is a byproduct obtained from cheese manufacturing. In the second experiment, using Dataset 2, we have performed binary classification in order to determine whether the samples are pure milk or modified milk. In Figure~\ref{fig:instances_b_a}-(B) we show IRT difficulty and discrimination of this dataset samples, and from these values distribution, we can see that the instance difficulties in this case are similar to the adulterants dataset, but discriminative samples are more equally distributed.

The third experiment uses the Trees Dataset, which consists of spectroscopy readings (spectral curves) from different tree organs such as roots, stems, branches, and leaves, obtained from 73 tree species in tropical and temperate biomes, from various forest types. This dataset is publicly available \citep{Ramirez2015}. In this experiment, we have performed classification considering 4 classes, in order to determine which of the tree organs analyzed the samples belong to: root, stem, branch, and leaf. IRT parameters from these samples, presented in Figure~\ref{fig:instances_b_a}-(C), show that there are relatively more difficult instances than the other problems, and most instances have low discriminative value.

\begin{figure}[ht!]
\centerline{\includegraphics[width=1\textwidth]{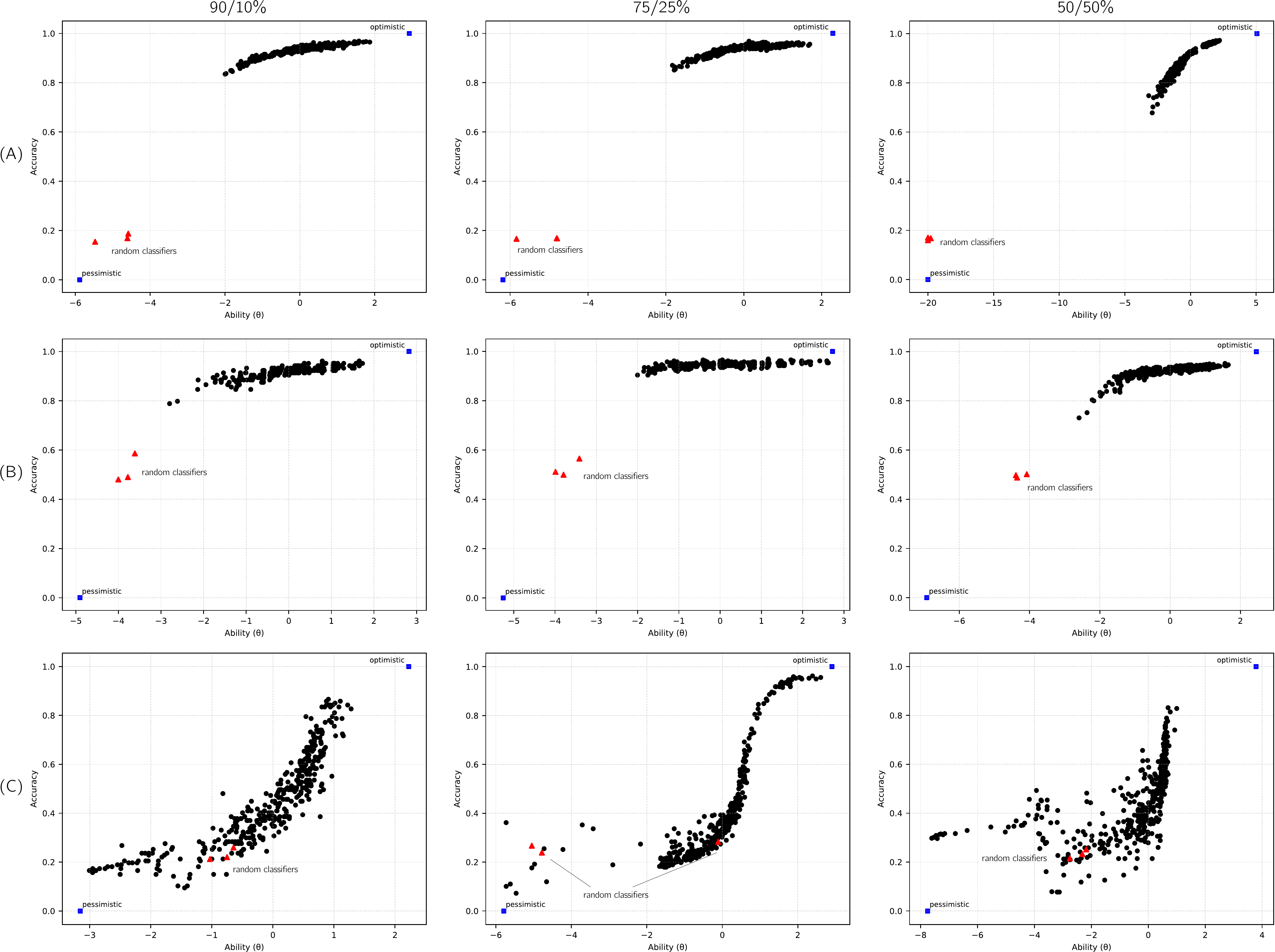}}
\caption{Scatter plot showing the relationship between the ability parameter ($\theta$) and the classifier accuracy for each dataset: (A) Milk Adulterants, (B) Milk Whey, (C) Trees; and for each training and test folds (90/10\%, 75/25\%, and 50/50\%). Optimistic and pessimistic classifiers are represented with squares, and random classifiers are represented with triangles. In general, the higher ability a model has, the higher its classification accuracy rate.}
\label{fig:ability_accuracy}
\end{figure}

As suggested by the work of \citeonline{MartinezPlumed2019}, we show the relationship between the ability estimated by the IRT model and the classifier accuracy for each dataset experimented. Figure~\ref{fig:ability_accuracy} shows that relationship for each dataset and for each training and test fold. The rows show the relationship from each dataset: (A) Milk Adulterants, (B) Milk Whey, and (C) Trees. Each column represents the training and test set folds: 90/10\%, 75/25\%, and 50/50\%. Figure~\ref{fig:ability_accuracy} also shows the artificial classifiers: optimistic and pessimistic at the borders of the plots, and random classifiers that produce accuracy below the 0.20 level in row (A), around 0.50 in row (B) and around 0.25 in row (C). The behavior of random classifiers is expected. They produce equally distributed sample predictions, so in (A) their accuracy is around 16\% because the dataset has 6 possible classes ($1 / 6 \approx 0.1667$), in (B) the accuracy is around 50\% because it is a binary classification, and in (C) the accuracy is around 25\% because there are 4 possible classes.

Figure~\ref{fig:ability_accuracy} also shows that, in general, the higher the ability estimated through IRT, the more correct predictions the trained model will produce. Also, the lower the training set ratio, the more models with lower abilities will be present. Finally, we can see that the problems in the different datasets have distinct inherent difficulties. In Figure~\ref{fig:ability_accuracy}-(A), the models are more condensed on higher accuracy levels and abilities around -2 and 2 in 90/10\% and 75/25\% folds; in 50/50\% fold, ability levels tend to have a higher impact on accuracy. In Figure~\ref{fig:ability_accuracy}-(B), the models are less condensed and their ability ranges are higher in the first two folds, and similar in the last fold to the previous dataset. In Figure~\ref{fig:ability_accuracy}-(C) the models are much more scattered, meaning that lower ability models have much more impact on accuracy than the other datasets.

\subsubsection*{Experiment results}

\begin{figure}[ht!]
\centerline{\includegraphics[width=1\textwidth]{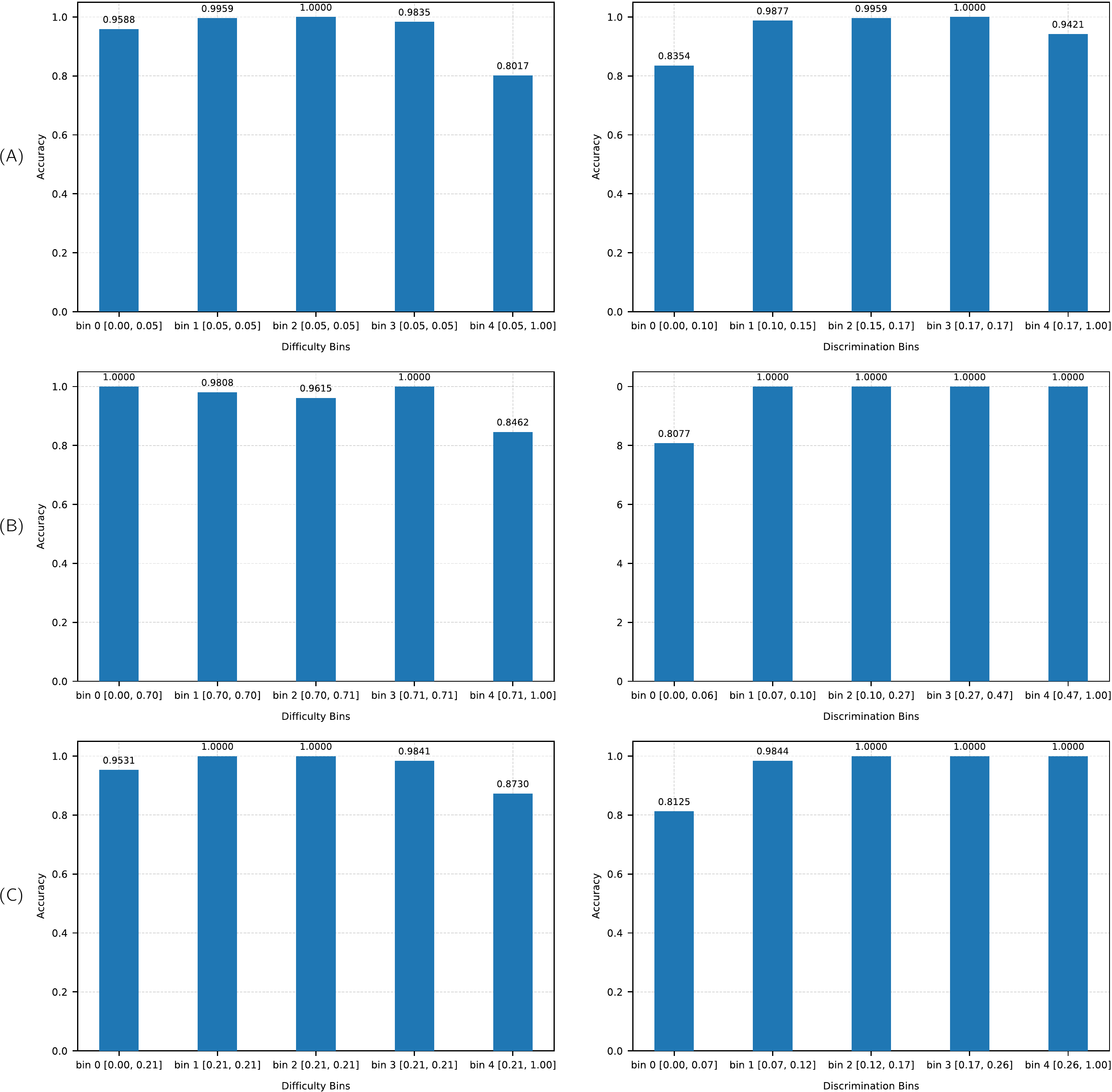}}
\caption{Bar plot of the proposed meta-learning classification considering difficulty and discrimination bins using the 75\%/25\% training/test fold for datasets: (A) Milk Adulterants, (B) Whey, and (C) Trees. Each bar represents the classification accuracy (y-axis) of a model that classifies only instances within the difficulty or discrimination range associated with the bar (described in x-axis). On top of each bar, there is the exact accuracy from that specific model.}
\label{fig:bins}
\end{figure}

\hl{The proposed methodology has been applied to each dataset three times using bootstrap sampling, for each training and test fold defined, which generated a visualization of the models' executions in each difficulty or discrimination bin. One method execution for the 75/25\% fold is shown} in Figure~\ref{fig:bins}, where rows (A), (B), and (C) represents the datasets used, and  the two columns show the histograms for difficulty and discrimination parameters, respectively. In the histograms, in each bar we can see the performance of a model that has the ability level compatible to the instances belonging to that bin. Considering difficulty bins, we can see that the performance decreases with the execution of the last bin, which is expected since it is the bin with the most difficult instances. Some lower accuracies are also found in the easiest bin (for instance, $\approx0.95\%$ with Milk Adulterants and Trees datasets). In either case, the remaining bins present very high accuracy values. Considering discrimination, the first bin in all datasets present lower accuracies, just above $0.80\%$, and all the other bins present very high accuracy values.

The experiments were performed considering both difficulty and discrimination bins though all steps defined in our methodology, but we also performed testing of $n$ models in the defined IRT rank and accounted correct predictions using majority-voting counting. Results with high accuracy in the voting indicates a good selection of models in the rank, made through IRT parameters. Since $n$ is small, the voting performed with only $n$ models is preferred over all trained models. So, three different results compose our AutoML results: methodology steps with difficulty bins and discrimination bins, and majority-voting with models from the rank. These results were compared to results from other methods as our benchmarks: a single CNN model, a general majority voting approach, and an Auto-Keras execution.

The single CNN model has a relatively simple architecture that has previously achieved great performance on spectral data \citep{AsseissNeto2019}, with one convolutional layer that learns 32 filters of kernel size 5, and a 1024 neurons fully connected layer. LeakyReLU activations to add non-linearity to the model \citep{Maas2013}, and dropout operations were used to prevent model overfitting \citep{Srivastava2014}. The voting approach obtained results from classifications with all models trained in the initial CNN collection, and the final prediction for each instance was the most frequent prediction within all models. Auto-Keras was performed with default configurations for searching model architectures using the ``AutoModel'' feature, which trained CNN models with variations of the following hyperparameters: dropout rate, learning rate, optimizer, number of convolution and dense layers, convolution kernel size, number of nodes of each layer, and the use or not of batch normalization. Auto-Keras works by running different ``trials'' of neural architectures and it uses the technique of architecture morphing \citep{Jin2019} in order to obtain incremental modifications of the networks. In our experiments, the number of trials to run Auto-Keras was set to 384, the same number of models that our methodology generated.

Tables~{\ref{tab:5_adulterants_accuracy}}, {\ref{tab:whey_accuracy}}, and {\ref{tab:trees_accuracy}} show accuracy comparison for the Milk Adulterants Dataset, the Whey Dataset, the Trees Dataset, respectively. These tables present, in the first three columns, \hl{average accuracy considering all three bootstrap samples for} executions that are part of our methodology: IRT difficulty and IRT discrimination columns refer to the steps proposed in Section~{\ref{subsec:proposed_methodology}} using difficulty and discrimination parameters, while the third column refers to a majority-voting with models from the IRT rank defined. The last three columns of the tables refer to different methods used as benchmark: the single CNN model, the voting approach considering all models generated in the CNN collection, and an Auto-Keras execution. \hl{Results for each individual bootstrap sampling are shown in Supplementary Material.}

\begin{table}[ht!]
\centering
\caption{Comparison of accuracy results for the Milk Adulterants Dataset considering the proposed methodology and benchmark methods.}
\label{tab:5_adulterants_accuracy}
\resizebox{\textwidth}{!}{
\begin{threeparttable}
\begin{tabular}{cccc@{\hskip 20pt}ccc}
\toprule

  & \multicolumn{3}{c}{Proposed AutoML Methodology} & \multicolumn{3}{c}{Benchmark} \\ \cmidrule(l{0pt}r{10pt}){2-4} \cmidrule(l{0pt}r{0pt}){5-7}

\thead{Fold}  & \thead{IRT\\difficulty} & \thead{IRT\\discrimination} & \thead{Voting\\(IRT rank)} & \thead{Single\\CNN model} & \thead{Voting\\(all models)} & \thead{Auto-Keras} \\ 
\midrule
90/10\% &   \hl{0.9772}	&   \hl{0.9779}	&   \hl{0.9766}	&   0.9608	&   0.9526	&	0.9567\\

75/25\% &   \hl{0.9713}	&   \hl{0.9705}	&   \hl{0.9749}	&   0.9685	&   0.9554	&	0.9537\\

50/50\% &   \hl{0.9734}	&   \hl{0.9733}	&   \hl{0.9668}	&   0.9538	&   0.9596	&	0.9442\vspace{3pt}\\

Average &   \hl{0.9740}	&   \hl{0.9739}	&   \hl{0.9728}	&   0.9610	&   0.9558	&	0.9515\\
\bottomrule
\end{tabular}

%\begin{tablenotes}[flushleft]\scriptsize
%\item \hspace{-5pt} The proposed methodology was performed following the steps for IRT bins, and also testing instances and accounting for classification using voting for models in the IRT rank. The benchmark is a single CNN model, a voting approach considering all models generated in the CNN collection, and an Auto-Keras execution.
%\end{tablenotes}

\end{threeparttable}
}
\end{table}

\begin{table}[ht!]
\centering
\caption{Comparison of accuracy results for the Milk Whey Dataset considering the proposed methodology and benchmark methods.}
\label{tab:whey_accuracy}
\resizebox{\textwidth}{!}{
\begin{threeparttable}
\begin{tabular}{cccc@{\hskip 20pt}ccc}
\toprule

  & \multicolumn{3}{c}{Proposed AutoML Methodology} & \multicolumn{3}{c}{Benchmark} \\ \cmidrule(l{0pt}r{10pt}){2-4} \cmidrule(l{0pt}r{0pt}){5-7}

\thead{Fold}  & \thead{IRT\\difficulty} & \thead{IRT\\discrimination} & \thead{Voting\\(IRT rank)} & \thead{Single\\CNN model} & \thead{Voting\\(all models)} & \thead{Auto-Keras} \\ 
\midrule
90/10\% &  \hl{0.9455}  &	\hl{0.9519}  &	\hl{0.9455}  &	0.9231  &	0.9231	&	0.9038\\

75/25\% &  \hl{0.9448}  &	\hl{0.9448}  &	\hl{0.9397}  &	0.9462  &	0.9500	&	0.9423\\

50/50\% &  \hl{0.9339}  &	\hl{0.9313}  &	\hl{0.9294}  &	0.9365  &	0.9346	&	0.9230\vspace{3pt}\\

Average &  \hl{0.9414}  &	\hl{0.9427}  &	\hl{0.9382}  &	0.9353  &	0.9359	&	0.9230\\
\bottomrule
\end{tabular}

\end{threeparttable}
}
\end{table}

\begin{table}[ht!]
\centering
\caption{Comparison of accuracy results for the Trees Dataset considering the proposed methodology and benchmark methods.}
\label{tab:trees_accuracy}
\resizebox{\textwidth}{!}{
\begin{threeparttable}
\begin{tabular}{cccc@{\hskip 20pt}ccc}
\toprule

  & \multicolumn{3}{c}{Proposed AutoML Methodology} & \multicolumn{3}{c}{Benchmark} \\ \cmidrule(l{0pt}r{10pt}){2-4} \cmidrule(l{0pt}r{0pt}){5-7}

\thead{Fold}  & \thead{IRT\\difficulty} & \thead{IRT\\discrimination} & \thead{Voting\\(IRT rank)} & \thead{Single\\CNN model} & \thead{Voting\\(all models)} & \thead{Auto-Keras} \\ 
\midrule
90/10\% &  \hl{0.9842}  &	\hl{0.9842}  &	\hl{0.9790}  &	0.9764  &	0.9843	&	0.9527\\

75/25\% &  \hl{0.9664}  &	\hl{0.9685}  &	\hl{0.9643}  &	0.9340  &	0.7701	&	0.8993\\

50/50\% &  \hl{0.9265}  &	\hl{0.9327}  &	\hl{0.7722}  &	0.4425  &	0.3176	&	0.9259\vspace{3pt}\\

Average &  \hl{0.9590}  &	\hl{0.9618}  &	\hl{0.9052}  &	0.7843  &	0.6906	&	0.9181\\
\bottomrule
\end{tabular}

\end{threeparttable}
}
\end{table}

Accuracy values show that the proposed AutoML methodology can outperform the benchmarks in most cases. For the Milk Adulterants Dataset (Table~\ref{tab:5_adulterants_accuracy}), on average, all methods presented approximately the same performance, with accuracies around 96\%, but the accuracies from our methodology are slightly higher than the benchmark ones. Our results are specifically better for the 50/50\% fold, which indicates that the methodology is less sensitive to a lower size of training dataset. For the Whey Dataset (Table~\ref{tab:whey_accuracy}) the proposed methodology considering both difficulty and discrimination presented the best performance on average. Voting with models from the rank had intermediate accuracy values, followed by the benchmark methods. Lastly, for the Trees Dataset (Table~\ref{tab:trees_accuracy}), all methods presented the same behavior, producing better results with higher training set. Benchmark methods, however, are clearly more sensitive to training size. \hl{In the Trees Dataset (Table~{\ref{tab:trees_accuracy}}), considering the 50/50\% fold, accuracy values drop sharply for the single CNN model and for voting with all models. This is because the single model was manually designed previously for other datasets. Also, the voting considers all trained models, which includes models with lower abilities. This is also shown in Figure~{\ref{fig:ability_accuracy}}-(C), that shows all models' abilities. Finally,} comparing both voting methods, the results highlight that the IRT selection of models not only drastically reduces the number of models considered for the majority-voting, but also offers greater performance.

\subsubsection*{Model complexities}

In order to give an overview of the model complexities that were part of the experiments, we calculate the number of parameters of each network, which generally can be considered as weights that are adjusted (learnt) during training. These weights contribute to the model's predictive power and the total count of parameters can be considered as a complexity of the model.

In our experiments, the proposed AutoML method is composed by a rank of CNN models that achieves the highest IRT scores. The number of models in this rank is arbitrary and suggested to be very low (i.e., 5). Then, we can see the complexity of the NASirt method as the sum of the number of parameters (weights) in each neural network. As for the benchmark methods, we explored a single CNN model, a voting method for all trained models, and the use of Auto-Keras. In the first benchmark, the complexity is simply the number of network parameters of the single model; for the second benchmark, the complexity is the sum of network parameters of all individual models. Auto-Keras, on the other hand, uses a morphism technique for generating neural architectures {\citep{Jin2019}}, therefore creating final models with lower number of parameters. Network parameter counting are presented in Table~{\ref{tab:network_params}}, where it is shown a quantitative comparison of complexities for the proposed and the benchmark methods. The table shows that the IRT rank selection produces overall complexities closer to the benchmark single model and Auto-Keras, which are much lower than the voting with all models.

% Therefore, in combination with the results shown in Tables~{\ref{tab:5_adulterants_accuracy}}, {\ref{tab:whey_accuracy}}, and {\ref{tab:trees_accuracy}}, we can say that the use of our meta-learning method is preferred over other methods.

\begin{table}[ht!]
\centering
\caption{Number of network parameters for models in the methodology rank, for the single CNN model, for the voting with all trained models, and Auto-Keras used as benchmark.}
\label{tab:network_params}
%\resizebox{\textwidth}{!}{
\begin{threeparttable}
\begin{tabular}{crrrr}
\toprule
\thead{Dataset}  & \thead{IRT Rank} & \thead{Single CNN} & \thead{Voting} & \thead{Auto-Keras} \\ 
\midrule
Milk Adulterants		&	46.8M		&	30.6M	&	3.7B		&	20.1M \\
Milk Whey			&	107.4M		&	30.6M	&	3.7B		&	20.9M \\
Trees				&	346.7M		&	69.7M	&	8.2B		&	72.1M \\
\bottomrule
\end{tabular}

\begin{tablenotes}[flushleft]\scriptsize
\item \hspace{-6pt} Values refer to the highest number of parameter among all training/test set folds, for each dataset.
\end{tablenotes}

\end{threeparttable}
%}
\end{table}

It is important to highlight that, although Auto-Keras delivers good performance with lower parameter complexities, the proposed NASirt method advances more towards the Explainable Artificial Intelligence (XAI). NASirt allows one to explain models' decisions locally using the IRT model and its difficulty and discrimination parameters. Thus, NASirt can be considered a more ``transparent'' AutoML method than Auto-Keras.

\section{Conclusion and future work}
\label{sec:conclusion}

In this work, we have explored a new AutoML classification approach for spectral datasets that uses IRT in order to estimate important characteristics on an instance-level. The proposed approach searches for high ability CNN models and it can automatically determine good hyperparameter combinations for neural networks. Our method relies on important information provided by IRT, like instance difficulty and discrimination, and inherit abilities of models.

We have performed several experiments with different spectral datasets in order to demonstrate the performance of the proposed methodology in different scenarios. Our methodology determines a specific CNN model to deal with groups (bins) of instances with similar characteristics, and classification is performed specifically for each bin. Thus, each classification from our method can provide a histogram that shows performance for each bin. It is shown, for instance, that in most cases, several bins had 100\% accuracy on their instance classifications. Later, those classifications are combined to generate the method's final prediction. Moreover, the ability rank determined through the use of IRT can reduce the number of models of a general meta-learning method that combines results with majority-voting, and yet it provides better accuracy than the indiscriminate use of model combinations.

Currently, the work described in this paper has the opportunity of improvements. So, as part of future works, we intent to use different AutoML performance estimation strategies, such as lower fidelity training, in order to speed-up initial model training and get ability estimates beforehand \citep{Elsken2019}. Also, there are opportunities to make the network architectures even more flexible with the use of more hyperparameter combinations. Finally, since it has been demonstrated that finding hyperparameters with random search can outperform grid search \citep{Bergstra2012}, we can implement some form of random search in the model generation step of our method.

\subsubsection*{\footnotesize Supplementary Material}

\footnotesize
Supplementary material such as dataset samples, additional figures, and the Python code needed to run NASirt are available at OSF: \url{https://osf.io/mrqc3/}.

\bibliography{manuscript}

\end{document}